\def\year{2019}\relax
\documentclass[letterpaper]{article} %DO NOT CHANGE THIS
\usepackage{aaai19}  %Required
\usepackage{times}  %Required
\usepackage{helvet}  %Required
\usepackage{courier}  %Required
\usepackage{url}  %Required
\usepackage{graphicx}  %Required
\usepackage{amsmath,amssymb,amsfonts}
\usepackage{natbib}
\usepackage{xfrac}
\usepackage{fancyhdr}
\nocopyright{}

\begin{document}
\title{Discretizing Logged Interaction Data Biases Learning for Decision-Making}
\author{Peter Schulam\\
Department of Computer Science\\
Johns Hopkins University\\
\texttt{pschulam@cs.jhu.edu}
\And
Suchi Saria \\
Department of Computer Science \\
Johns Hopkins University\\
\texttt{ssaria@cs.jhu.edu}
}
\maketitle

\begin{abstract}
  Time series data that are not measured at regular intervals are commonly \emph{discretized} as a preprocessing step. For example, data about customer arrival times might be simplified by summing the number of arrivals within hourly intervals, which produces a discrete-time time series that is easier to model. In this abstract, we show that discretization introduces a bias that affects models trained for decision-making. We refer to this phenomenon as \emph{discretization bias}, and show that we can avoid it by using continuous-time models instead.
\end{abstract}

\section{Introduction}

Time series data sets are often used to train models for decision-making. For example, website logs recording the ads served to users and whether they clicked-through can help to build algorithms that make better choices for future visitors. In medicine, electronic health record data containing patient treatment histories and laboratory test results can inform which treatments to give to new patients and when. We refer to this type of data as \emph{logged interaction} data.

Decision-making problems are often formalized using discrete-time models such as a discrete-time Markov decision process (D-MDP). Although D-MDPs are formulated in discrete-time and assume observations lie on a regular grid, time series data does not often satisfy this assumption. For example, users do not visit sites and click ads on a regular discrete-time grid. Similarly, a doctor does not make treatment decisions and choose to order laboratory test results in discrete-time, but rather chooses when to treat and when to measure at points in continuous-time based on her evolving understanding of the patient's condition. Time series data sets like these examples are comprised of irregularly spaced observations. Even if the data is generated from a discrete-time system, missingness can cause the data to appear irregularly spaced. To model such data, practitioners discretize the irregularly spaced observations into equal-sized time windows (e.g. by binning the data and computing averages within each bin).
% To preprocess such data so that they can apply discrete-time models, practitioners will often first \emph{discretize} irregularly spaced time series data by partitioning the values into equally-sized time windows, and then aggregating within each bucket (e.g. by averaging). 

In this paper, we show that discretization can bias predictive models for decision-making, which can lead to incorrect or harmful downstream decisions (see e.g. \citealt{schulam2017reliable}). We refer to this phenomenon as \emph{discretization bias}. At a high level, discretization bias is a form of \emph{confounding}, and so is related to biases that are adjusted for in off-policy reinforcement learning using, for example, propensity score weights (e.g. \citealt{swaminathan2015counterfactual}). There is not, however, any work discussing how common preprocessing steps can introduce such biases. As a solution, we show that continuous-time models do not suffer from discretization bias and may therefore be more reliable for solving sequential decision problems.

% Actions in time series data are often chosen dynamically based on the history of the time series. For example, treatments that a doctor chooses may be driven by the latest laboratory test results and when they were taken. The history is therefore a confounder for models that predict future values of the time series. Discretization drops information from the history and induces confounding. As a solution, we show that continuous-time models do not suffer from discretization bias and are therefore more reliable for solving sequential decision problems.

\begin{figure*}[t]
  \centering
  \includegraphics[width=0.88\linewidth]{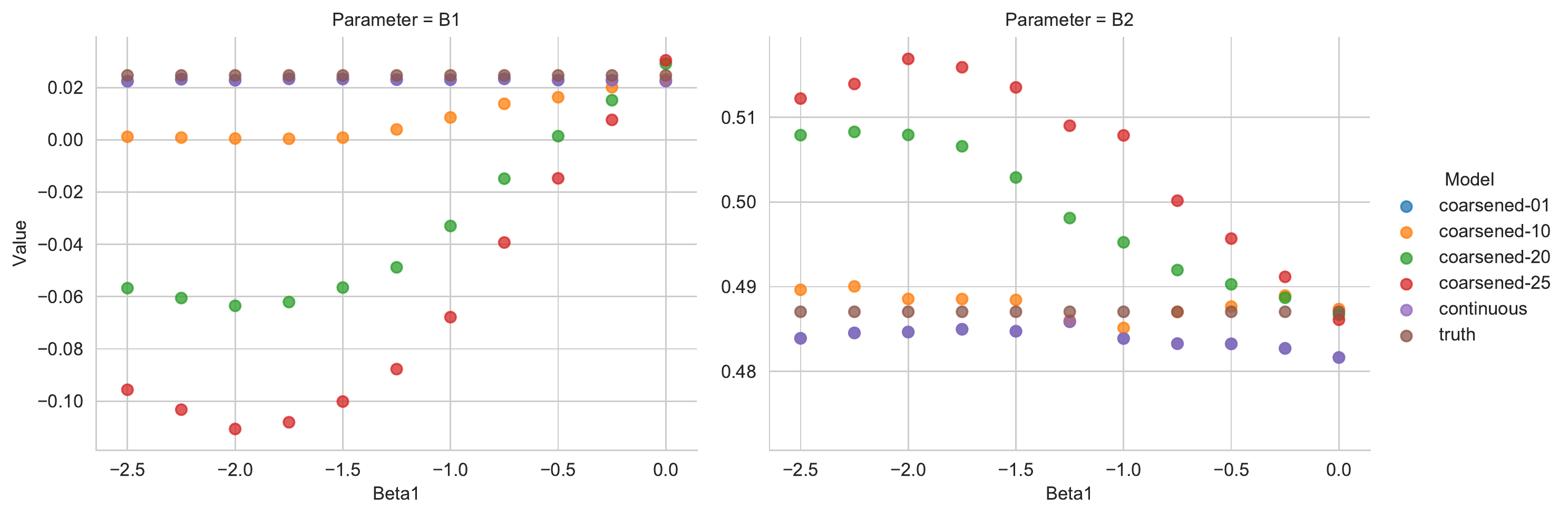}
  \caption{Action effect estimates (y-axis) for each model under policies with varying levels of dependence on the history (x-axis). The \texttt{coarsened-01} and \texttt{continuous} models produce the exact same estimates, and so are overlaid in the plots.}
  \label{fig:results}
\end{figure*}

\vspace{-12pt}
\paragraph{Outcome model.}
An important element of D-MDPs and other sequential decision-making frameworks is the \emph{outcome model}, which predicts future values of the time series given the history. Let $Y_k$ be the $k^{\text{th}}$ observation of a discrete-time time series and let $U_k$ denote the $k^{\text{th}}$ action, then
\begin{align}
  \label{eq:model}
  P \big( Y_k \mid \mathcal{Y}_k, \mathcal{U}_k \big)
\end{align}
is the outcome model, where $\mathcal{Y}_k$ is all previous observations $[Y_1, \ldots, Y_{k-1}]$ and $\mathcal{U}_k$ is all previous actions $[U_1, \ldots, U_{k-1}]$. The outcome model is important, for example, in off-policy reinforcement learning where the goal is to learn good action policies from retrospective time series data (e.g. \citealt{dudik2011doubly}). We assume that the distribution of actions $U_k$ is determined by a \emph{policy} $\pi$, that may depend on $\mathcal{Y}_{k}$, $\mathcal{U}_{k}$, and $Y_k$, which we call the \emph{history} and denote using $\mathcal{H}_k$.

\vspace{-12pt}
\paragraph{Causal inference and confounding.}
Equation \ref{eq:model} must be a \emph{causal model} that predicts how $Y_k$ would vary if we were to intervene and choose new actions $\mathcal{U}_k$. This is only a concern when learning from retrospective, or logged, interaction data (i.e. the algorithm cannot choose actions). One way to ensure that Equation \ref{eq:model} is a causal model is to check the backdoor criterion, which dictates that all confounders of the effect of $\mathcal{U}_k$ on $Y_k$ are included as predictors \citep{pearl2009causality}. A confounder is any variable that affects the distribution of $\mathcal{U}_k$ (i.e. it is used in the policy $\pi$) and also the distribution of $Y_k$. In time series data, the policy $\pi$ can depend on the history $\mathcal{H}_k$, and so elements of the history may confound the effect that $U_k$ has on future values $\{ Y_{k'} : k' > k \}$. Therefore, if an element of $\mathcal{H}_k$ that affects the policy $\pi$ is not included in the outcome model, the model may be biased.

\vspace{-12pt}
\paragraph{Discretization bias.}
Discretization bias is a form of confounding caused by \emph{grouping individual observations} into equally sized bins and creating an aggregate observation by summarizing those observations (e.g. using the average). This preprocessing step treats groups of observations as exchangeable, and drops information about their specific values and measurement times. If the dropped information is used in the policy, then Equation \ref{eq:model} may be biased.

\section{Simulation Experiment}

We now describe a simulation experiment that shows how discretization bias can cause a learning algorithm to recover the wrong outcome model parameters. We simulate time series data from a two-dimensional discrete-time Gaussian hidden Markov model (SG-HMM), where $\delta$ is the time between steps in the SG-HMM. The generative model is:
\begin{align}
  X_k \mid X_{k-1}, U_{k-1}
    &\sim \mathcal{N}\big( A X_{k-1} + B U_{k-1}, C \big) \\
  Y_k \mid X_k
    &\sim \mathcal{N}\big( H X_{k}, R \big).
\end{align}
The action variables $U_k \in \{0, 1\}$ indicate whether an action was taken at time $k$ (1 if an action was taken, 0 otherwise). The effect of the action on future values of the time series is determined by the matrix $B$. The specific SG-HMM that we use is derived by discretizing a continuous-time stochastic spring model. Details are provided in the supplement.

To mimic irregularly sampled data, each observation $Y_k$ is observed with probability $\sfrac{1}{5}$. The variable $O_k = 1$ if $Y_k$ is observed, and $O_k = 0$ otherwise.
% (i.e. the data is missing completely at random; \citealt{little2014statistical}).
The action $U_k$ has a distribution that depends on all previously observed values $\{ Y_{k'} : k' \leq k, O_{k'} = 1 \}$. The details of this distribution are in the supplement. Intuitively, the distribution is controlled by a parameter $\beta_1$. When $\beta_1 < 0$, then actions are more likely when recent values of $Y_k$ are low. As $\beta_1 \to 0$, the actions are chosen independently of the history, and so confounding is no longer an issue.

To discretize the simulated data, we define the \emph{coarsening factor} $m$. We divide the time interval into windows of size $\Delta = m \delta$ and average all $Y_k$ where $O_k = 1$ in the same window to create the discretized time series. The actions $U_k$ are not averaged, but instead are concatenated into a length-$m$ vector. Discretizing $Y_k$ with coarsening factor $m$ is a linear operation, and so the discretized data is sampled from an SG-HMM with exactly the same parameters as the original data. We refer to this new SG-HMM as the \emph{coarsened model}. A description of how to construct the coarsened model for a given coarsening factor $m$ may be found in the supplement.

We fit five models using maximum likelihood: four discrete-time models with coarsening factors 1, 10, 20, and 25 (when $m = 1$, the coarsened model is the original SG-HMM), and one continuous-time model. Because the SG-HMM is derived by discretizing a continuous-time model, all five models depend on the same underlying parameters. Maximizing the likelihood of an SG-HMM is a non-convex optimization problem, so we simplify by assuming that all parameters are known \emph{except} for the matrix $B$. Estimating $B$ is a concave optimization problem, which simplifies estimation and helps to isolate the effects of discretization.

Figure \ref{fig:results} displays the results of the simulation experiment. We see that as $\beta_1 \to 0$, both elements of $B$ (listed as \texttt{B1} and \texttt{B2}) are accurately estimated for all models. As $\beta_1$ becomes more negative, however, we see that the models with larger coarsening factors $m$ are biased. In particular, note that the models with $m = 20$ and $m = 25$ learn that actions have the opposite effect (i.e. \texttt{B1} is negative instead of positive). On the other hand, the continuous-time model gives the exact same, unbiased, estimates of the parameters as the true model with coarsening factor $m = 1$.

\vspace{-14pt}
\paragraph{Discussion.}
We introduced the idea of \emph{discretization bias}, which affects discrete-time models learned for decision-making. If trained correctly, continuous-time models can avoid this bias. A set of conditions for correctly training such models are laid out in \citet{schulam2017reliable}: we cannot drop values or measurement times because the policy may depend on them. Discretization removes this information, and therefore may cause a predictive model to violate those conditions.

% We have introduced the idea of \emph{discretization bias}, which is caused by discretizing time series data before estimating a causal model. Using simulated data, we showed that discretization bias can have a significant impact on the accuracy of Equation \ref{eq:model}. Our experiments showed that discretization can estimate that an action has the opposite effect on future outcomes. On the other hand, we showed that continuous-time models, which do not depend on discretization as a preprocessing step, do not have this bias. We conclude that continuous-time models may be more reliable for modeling the effects of actions in sequential decision problems.

\vspace{-6pt}
\bibliographystyle{aaai}
{\small \bibliography{main}}

\appendix

\section{Supplement}

%% \section{Causal Inference and Confounders}

%% Learning the effect of an action on an outcome has been formalized in the field of causal inference (see e.g. \citealt{pearl2009causality}). In this paper, we focus on an approach that we refer to as the \emph{direct approach}. In the direct approach, we estimate the conditional distribution of the outcome given an action. The key difficulty in the direct approach is to ensure that the conditional distribution is not biased by \emph{confounders}. For example, suppose we find that $\sfrac{1}{4}$ of people who take drug $A$ die, but only $\sfrac{1}{8}$ of people who do not take the drug die. One conclusion is that $A$ is harmful. An alternative explanation is that those who take drug $A$ are sicker and at higher risk to begin with. In the latter explanation, the relationship is biased by the confounding variable \texttt{sickness}. Confounders are variables that affect the distributions of both the actions and outcomes in the training data. To remove the bias, we must include all confounders as additional predictors in the conditional distribution. Unfortunately, we cannot empirically check that all confounders are included. In practice, we must assume that we have included all confounders, which is known as the \emph{no unobserved confounders} assumption.

\subsection{Discrete-Time System}

We define a Gaussian discrete-time hidden Markov model by discretizing a stationary linear continuous-time hidden Markov model. A stationary linear continuous-time model hidden Markov model is parameterized by five matrices: $F$, $G$, $Q$, $H$, and $R$. The first three matrices describe the system \emph{dynamics}, and the final two matrices describe the \emph{observation model}. The dynamics are characterized using the It\^{o} stochastic differential equation
\begin{align}
  d X(t) = F X(t) dt + G U(t) + L d\beta(t),
\end{align}
where $U(t)$ is an input process and $L$ is the Cholesky factorization of the positive definite covariance matrix $Q$ and $\beta(t)$ is Brownian motion. The observation model is a simple multivariate normal distribution:
\begin{align}
  Y(t) \mid X(t) \sim \mathcal{N} \big( H X(t), R \big).
\end{align}

A Gaussian discrete-time hidden Markov model is also parameterized by five matrices: $A$, $B$, $C$, $H$, and $R$. As before, the first three matrices define the dynamics model and the last two define the obsrvation model. For $k \in [0, \ldots, n_k]$, we define the distributions of the discrete-time random variables
\begin{align}
  X_k \mid X_{k-1} &\sim \mathcal{N}\big( A X_{k-1} + B U_{k-1}, C \big) \\
  Y_k \mid X_k &\sim \mathcal{N}\big( H X_k, R \big).
\end{align}

\begin{figure}[ht]
  \centering
  \includegraphics[width=0.8\linewidth]{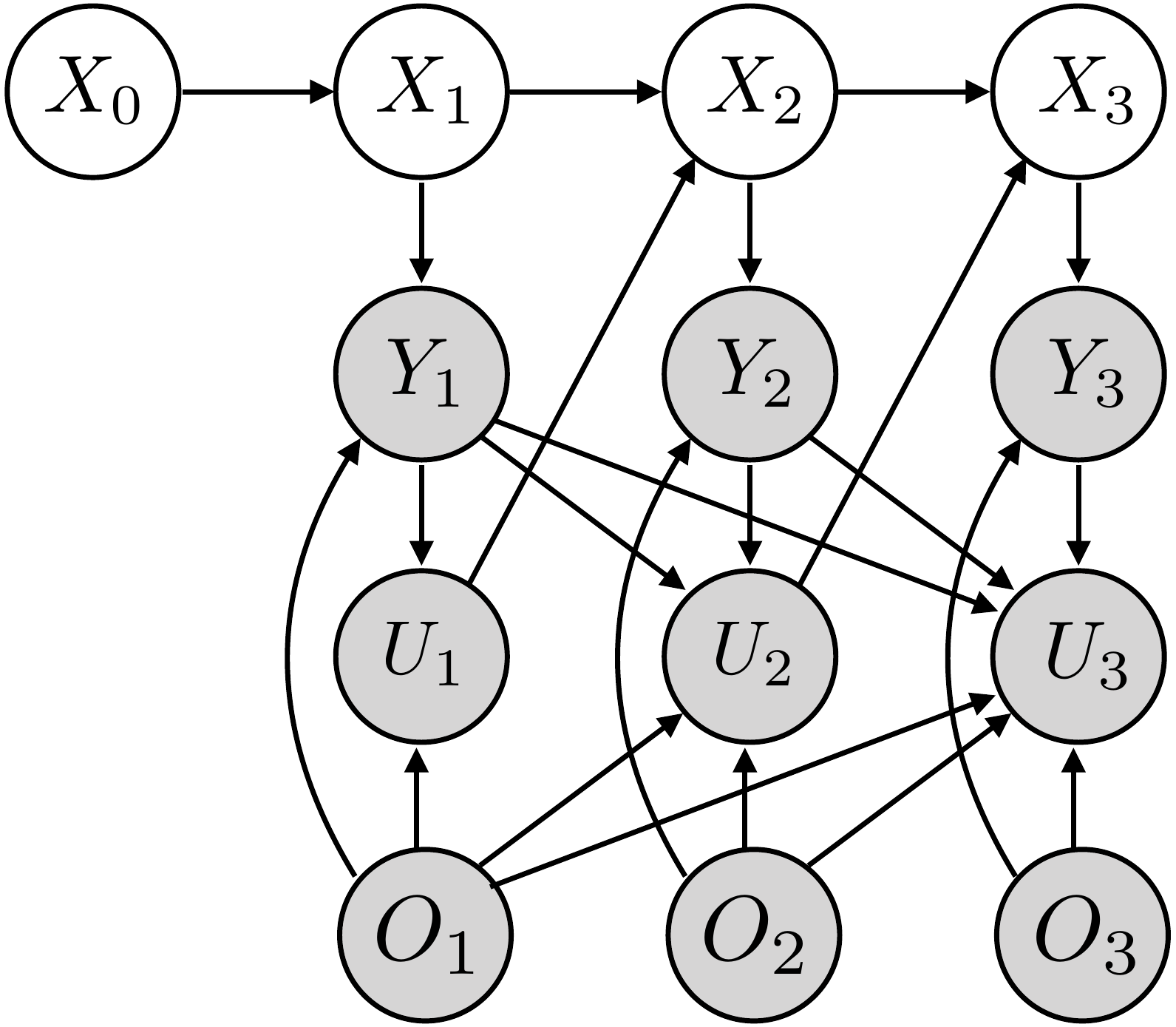}
  \caption{Graphical model for simulated data.}
\end{figure}

\subsubsection{Continuous to Discrete Conversion}

\begin{figure*}[t]
  \centering
  \includegraphics[width=0.9\linewidth]{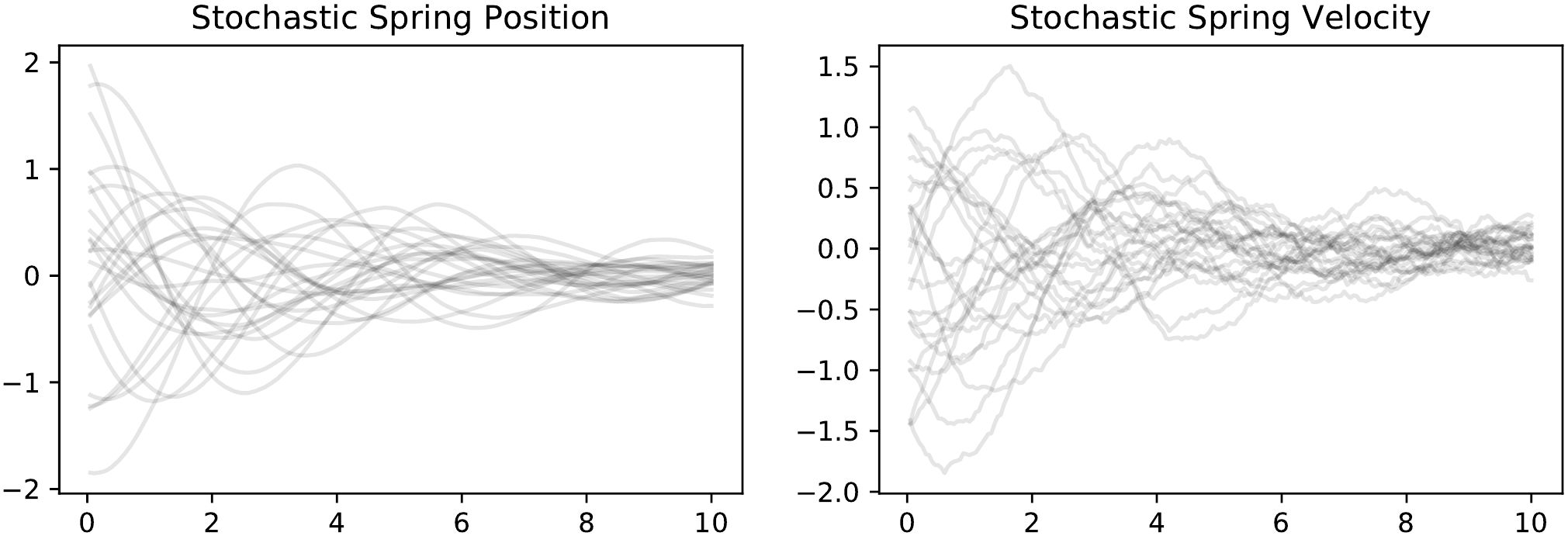}
  \caption{Sample trajectories from the stochastic spring model.}
  \label{fig:spring-samples}
\end{figure*}

To discretize the continuous-time model described above, we need to define a mapping from $(F, G, Q, H, R)$ to $(A, B, C, H, R)$. The mapping depends on the \emph{timestep} of the discretization, which we will denote using $\delta$. The timestep determines the intervals at which we observe the continuous-time process. Suppose that $X(t)$ is defined on the interval $[0, T]$ and that $\delta$ is chosen such that $n_k \delta = T$, then the discretization defines the random variables
\begin{align}
  X_k &\triangleq X(k \delta) \\
  Y_k &\triangleq Y(k \delta)
\end{align}
for $k \in [0, \ldots, n_k]$. Using these definitions, we first calculate the conditional distribution of $X_k$ given $X_{k-1}$. This conditional distribution has mean and covariance
\begin{align}
  \mu &= \Phi(\delta) X_{k-1} + \int_{0}^{\delta} \Phi(\delta - s) G U(k\delta - \delta + s) ds \\
  \Sigma &= \int_{0}^\delta \Phi(\delta - s) Q \Phi^T(\delta - s) ds,
\end{align}
where $\Phi(s) = \exp \{ F s \}$; the matrix exponential of $F s$ (see, e.g., \citealt{sarkka2014applied}). We therefore see that $A = \Phi(\delta)$ and that $C = \Sigma$ in our mapping from continuous-time to discrete-time. To define the matrix $B$, we make the assumption that $U(t)$ is defined on the grid $[0, \delta, 2\delta, \ldots, n_k \delta]$ using a sequence of $n_k + 1$ values $U_0, U_1, \ldots, U_k$:
\begin{align}
  U(t) = \sum_{i=0}^{n_k} U_i \delta_{i\delta}(t),
\end{align}
where $\delta_{i\delta}$ is the Dirac delta function centered at $i\delta$. If $U(t)$ has this form, then the integral in the expression for the conditional expected value $\mu$ of $X_k$ given $X_{k-1}$ is
\begin{align}
  \int_{0}^{\delta} \Phi(\delta - s) G U(k\delta - \delta + s) ds
  = \Phi(\delta) G U_{k-1}.
\end{align}
Therefore, we have $B = \Phi(\delta) G$. To complete the mapping, note that $H$ and $R$ do not need to be modified for the continuous to discrete conversion. In summary, we have
\begin{align}
  A &= \Phi(\delta) \\
  B &= \Phi(\delta) G \\
  C &= \int_{0}^\delta \Phi(\delta - s) Q \Phi^T(\delta - s) ds \\
  H &= H \\
  R &= R.
\end{align}

\subsubsection{Stochastic Spring Model}

For our experiments, we define the continuous-time model using a stochastic spring model. The dynamics of the stochastic spring depend on two parameters $\nu$ and $\gamma$, which we set to $1.0$ and $0.5$ respectively. The model is parameterized using
\begin{align}
  F &= \begin{bmatrix}
    0.0 & 1.0 \\
    -\nu^2 & -\gamma
  \end{bmatrix} \\
  G &= \begin{bmatrix}
    0.0 \\
    0.5
  \end{bmatrix} \\
  Q &= \begin{bmatrix}
    10^{-8} & 0.0 \\
    0.0 & 10^{-2}
  \end{bmatrix} \\
  H &= \begin{bmatrix}
    1.0 & 0.0
  \end{bmatrix} \\
  R &= \begin{bmatrix}
    10^{-4}
  \end{bmatrix}.
\end{align}

To simulate from the model, we draw an initial state $X(0)$ from a two-dimensional normal distribution with zero mean and identity covariance. Figure \ref{fig:spring-samples} shows a sample of simulated trajectories from the stochastic spring model.

\subsection{Simulating Actions}

The actions $U_k \in \{0, 1\}$ are chosen dynamically based on the history of observed measurements $[Y_1, \ldots, Y_k]$. In particular, each $U_k$ has a Bernoulli distribution with a mean parameter that is computed using a weighted average of the previously observed measurements. Let $\pi_k$ denote the expected value of $U_k$, then
\begin{align}
  \log \frac{\pi_k}{1 - \pi_k} = \beta_0 + \beta_1 \sum_{i=1}^k w_i Y_i.
\end{align}
To compute the weights $w_i$, we define a parameter $\alpha \in [0, 1]$. In a history of $k$ measurements, the weight for the $i^{\text{th}}$ measurement $Y_i$ is
\begin{align}
  w_i \propto \alpha^{k - i}.
\end{align}
The weights are normalized to sum to one. We see that when $\alpha = 0$, the history has no effect (i.e. the log odds are 0). On the other hand, when $\alpha = 1$ all of the previous measurements are weighted equally. When $\alpha \in (0, 1)$, the more recent measurements are given more weight.

\begin{figure}
  \centering
  \includegraphics[width=0.8\linewidth]{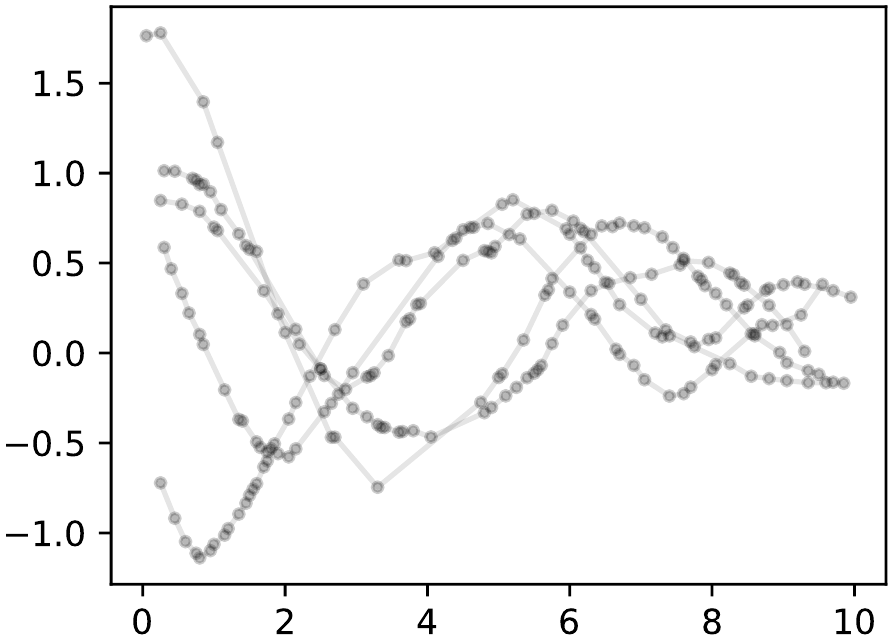}
  \caption{Examples of the data used to learn $B$.}
  \label{fig:treated-example}
\end{figure}

\paragraph{Missing data in histories.}
In our experiment, we randomly ``drop'' measurements $Y_k$ with probability $p_m = 0.8$. When a measurement is dropped, we replace it with the null value $\varnothing$. To account for missing data in the average used to compute $\pi_k$, we define the variable $O_k = 1$ when $Y_k$ is observed (i.e. it is not $\varnothing$), and $O_k = 0$ otherwise. We then modify the weight for measurement $Y_i$ to be
\begin{align}
  w_i \propto O_i \alpha^{k - i}.
\end{align}

\subsection{Coarsened Discrete Models}

When time series data is dicretized, the observations are first grouped into a sequence of equal-sized windows and then an aggregate measurement is computed from the measurements that fall into the bin. The average (arithmetic mean) is typically used as the aggregation method.

Recall that $\delta$ is the unknown step size of the discrete-time system that generated our data. To formalize the discretization preprocessing step, we introduce the idea of \emph{coarsening}, which is an operation on a discrete-time model that produces another discrete-time model with a timestep $\Delta$ that is larger than $\delta$. Coarsening depends on an integer $n_c \geq 2$, which we refer to as the \emph{factor}. Given the coarsening factor $n_c$, we define new states $X'_k$ and observations $Y'_k$ that are obtained by stacking $n_c$ consecutive states (or observations) together to form a larger vector:
\begin{align}
  X'_k &= \big[ X^T_{n_c(k-1) + 1}, \ldots, X^T_{n_c(k-1) + n_c} \big]^T \\
  Y'_k &= \big[ Y^T_{n_c(k-1) + 1}, \ldots, Y^T_{n_c(k-1) + n_c} \big]^T
\end{align}
If $(A, B, C, H, R)$ are the parameters of the original discrete-time HMM, then the distribution over the coarsened states $X'_k$ and $Y'_k$ is also a discrete-time HMM with parameters $(A', B', C', H', R')$ that depend only on $(A, B, C, H, R)$. The coarsened dynamics and measurement matrices have the following block structure:
\begin{align}
  [A']_{ij} &= \begin{cases}
    A^i & \text{ if $j = n_c$,} \\
    0   & \text{ otherwise.}
  \end{cases} \\
  [B']_{ij} &= \begin{cases}
    0 & \text{ if $i < j$,} \\
    A^{i-j} B & \text{ otherwise.}
  \end{cases} \\
  [C']_{ij} &= \begin{cases}
    C & \text{ if $i = j = 1$,} \\
    A^{i-1} C (A^{i-1})^T + C & \text{ if $i = j > 1$,} \\
    [C']_{ii} (A^{j-i})^T & \text{ if $i < j$,} \\
    A^{i-j} [C']_{jj} & \text{ if $i > j$.}
  \end{cases}
\end{align}
The coarsened measurement model parameters $H'$ and $R'$ also have block matrix structure:
\begin{align}
  [H']_{ij} &= \begin{cases}
    H & \text{ if $i = j$,} \\
    0 & \text{ otherwise.}
  \end{cases} \\
  [R']_{ij} &= \begin{cases}
    R &= \text{ if $i == j$,} \\
    0 &= \text{ otherwise.}
  \end{cases}
\end{align}

\subsubsection{Coarsening and Discretization}

To discretize data, there is typically an aggregation step that summarizes a collection of observations that fall into the same bin. One of the most common aggregation operations is taking the average of all observations in a bin, and this is how we aggregate in our simulation experiment. Since averaging is a linear operation, we see that preprocessing with discretization defines a new coarsened discrete-time HMM with a new measurement model
\begin{align}
  H'' &= [n_c^{-1}, \ldots, n_c^{-1}] \, H' \\
  R'' &= \sum_{i=1}^{n_c} n_c^{-2} R^2.
\end{align}

\end{document}